\documentclass{article}

\usepackage{graphicx}
\usepackage{subfigure}
\usepackage{makecell}
\usepackage{xcolor}
\usepackage{float}
\usepackage{booktabs}

\usepackage{hyperref}

\usepackage[accepted]{icml2020}

\begin{document}

\twocolumn[
\icmltitle{Weakness Analysis of Cyberspace Configuration \\ Based on 
Reinforcement 
Learning}
\icmlsetsymbol{equal}{*}
\begin{icmlauthorlist}
\icmlauthor{Lei Zhang}{}
\icmlauthor{Wei Bai}{}
\icmlauthor{Shize Guo}{}
\icmlauthor{Shiming Xia}{}
\icmlauthor{Hongmei Li}{}
\icmlauthor{Zhisong Pan}{}
\vspace{.1in}\\
Army Engineering University of PLA, Nanjing, China, 210007
\vspace{.1in}\\
zhangleicsphd@gmail.com, baiwei\_lgdx@126.com, szguo.uestc@gmail.com, 
pub\_xsm@hotmail.com, amayli003@sina.com, hotpzs@hotmail.com
\vspace{.1in}\\
\end{icmlauthorlist}

\vskip 0.3in
]



\printAffiliationsAndNotice{\icmlEqualContribution} 
\begin{abstract}
 In this work, we present a learning-based approach to analysis cyberspace 
 configuration. Unlike prior methods, our approach
has the ability to learn from past experience and improve over time. In 
particular, as we train over a greater number of agents as attackers, our 
method becomes better at rapidly finding attack paths for previously hidden 
paths, especially in multiple domain cyberspace. To achieve these results, we 
pose finding attack paths as a Reinforcement Learning (RL) problem and train an 
agent to find multiple domain attack paths. To enable our RL policy to find 
more hidden attack paths, we ground representation introduction an multiple 
domain action select module in RL. By designing a simulated cyberspace 
experimental environment to verify our method. Our objective is to find more 
hidden attack paths, to analysis the weakness of cyberspace configuration. The 
experimental results show that our method can find more hidden multiple domain 
attack paths than existing baselines methods.


\end{abstract}

\section{Introduction}

Rapid progress in AI has been enabled by remarkable advances in computer 
systems and hardware, but it is not widely used in cyberspace security 
protection. However, the intellectualization of cyberspace security protection 
system is an important problem facing the current cyberspace security 
protection\cite{DBLPSS10}. In the security management of cyberspace, actually 
the cyberspace should be regarded as a space composed of physical domain, 
digital domain and social domain, and its security protection should also be 
conducted as a whole\cite{DBLPnips2016}. In this process, it mainly includes 
the configuration of intelligence discovery of weakness, intelligent deployment 
of security equipment\cite{DBLPtsmc05}, intelligent monitoring of network 
traffic, intelligent awareness of security situation and other 
parts\cite{inbook009}, which comprehensively constitute the cyberspace security 
protection system. We believe that it is AI itself that will provide the means 
to constitute the cyberspace security protection system, creating a symbiotic 
relationship between cyberspace security and AI with each fueling advances in 
the other.

In this work, we present a learning-based approach to analysis cyberspace 
configuration. Our objective is to find more hidden attack paths, to analysis 
the weakness of cyberspace configuration. Despite of research on this problem, 
it is still necessary for human experts to realize the security risks existing 
in the current cyberspace of accurate judgment and evaluation, in order to 
ensure the security of the entire cyberspace. The problem’s complexity arises 
from multiple domain interaction each other in cyberspace, and only digital 
domain or network domain can be considered in current research. Even after 
breaking the problem into more manageable sub-problems, the state space is 
still orders of magnitude larger than recent problems on which 
intelligent-based methods have shown success.

To address this challenge, we pose find multiple domain attack paths as a 
Reinforcement Learning (RL) problem, where we train an agent (as an attacker) 
to find the attack paths. In each iteration of training, all of the attack 
paths are sequentially found by the RL agent. Training is guided by a fast but 
approximate reward signal for each of the agent’s find attack paths.


To our knowledge, in order to realize the target, the following problems need 
to be solved:

\begin{itemize}
	
	\item{First, the problem of multiple domain which can interaction each 
	other. This paper studies the problem of multiple domain cyberspace, and 
	changes the existing cyberspace security risk analysis to focus on multiple 
	domain cyberspace, enforce the pertinence and relevance of business level.}
	
	\item{Second, the problem of the alternative actions in different states is 
	different. In traditional RL, the number of actions which agent can select 
	is the same in any state. Therefore, this paper introduction the 
	multi-domain action select module in RL algorithm, in order to make the 
	alternative actions in different states is different.}
	
	\item{Third, the problem is how to measure the cyberspace weakness by 
	multiple domain cyberspace attack paths. This paper proposed an basic index 
	measured by the average attack paths of all attackers in the cyberspace. In 
	this way, we can measure the weakness of different cyberspace 
	configuration, thus provides the correlation reference for the different 
	cyberspace configuration, to help the administrator to improve the 
	cyberspace security.}
	
\end{itemize}

We believe that the ability of our approach to learn from experience and 
improve over time unlocks new possibilities for network administrator. We show 
that we can achieve superior result on simulated cyberspace experimental 
environment, as compared to the baselines method. Furthermore, our methods can 
find more hidden attack paths comparable to human expert based method in same 
time. Although we evaluate primarily on cyberspace configuration analysis, our 
proposed method is broadly applicable to many cyberspace security analysis.
 
\section{Related Work}
\subsection{Intelligent Security Protection}

Intelligent security protection mainly studied the behavior characteristics and 
rules of users network monitoring and network optimization. In general, there 
is an attack in the cyberspace, when it appears, the traffic will change, we 
can take advantage of the attack mode type to detect cyberspace anomalies. 
Collect the original message of the data in the network and extract it. Take 
the destination address and other information, establish the normal traffic 
model, and then use discrete wavelet transform technology analyzes and detects 
the data flow to judge the cyberspace anomalies\cite{TNET902685} . 

At present, the intelligent security protection technology based on cyberspace 
user's action mainly relies on web data mining, user abnormal action detection 
and neural network based method to distinguish.

Combining traditional data mining techniques with the Internet for web mining 
is to extract potentially useful patterns and hidden information from web 
documents, web structures and service logs. Generally, according to the 
different objects of web mining, people divide web data mining into three 
types: web content mining, web structures mining, use record mining and web 
comprehensive mining\cite{Badea2015}.

In the operation process of users will retain a lot of action information, 
effective use of this information is the basis and key to the realization of 
abnormal action determination. Multi-layer log collection is implemented to 
support the decision of user access action. Using multi-level user access log, 
and integrate web front end user click action and URL access logic, to extract 
the user's access action characteristics, by a large number of calculating the 
average user action baseline characteristics, use of effective monitoring 
abnormal access action scoring algorithm, trace the action of the abnormal IP, 
corresponding treatment measures\cite{Beutel2015}.

As an important method to deal with nonlinear systems, the neural network 
method has been successfully applied in the fields of pattern recognition and 
probability density estimation. Compared with the statistical analysis theory, 
the abnormal behavior analysis method based on neural network can better 
express the nonlinear relationship between one variable and another. The 
changing of abnormal network action requires the ability of behavior analysis 
system to analyze a large number of network packets. Moreover, many common 
attacks may be coordinated by multiple attackers on the cyberspace, which 
requires the network abnormal action analysis system must have the ability to 
deal with a large amount of nonlinear data. The method based on neural network 
has a fast response ability, especially for the processing of noisy data and 
incomplete data, so it provides a great flexibility for the analysis of 
intelligent security protection\cite{Kawazu2016Analytical}.


In recent years, the emergence of machine learning has made intelligent 
security protection become a new trend. There are many new attempts, including 
SVM\cite{DBLP123}\cite{Gao2017A}, K-nearest neighbors\cite{Xu2017Incremental}, 
Naive Bayes\cite{DBLPBK19}, random forests\cite{DBLPZH08}, neural 
network\cite{DBLPMK17}, deep learning and so on. The methods based on deep 
learning have become mainstream in the field because of their better 
performance. Gao proposed an model based on deep belief network, which uses a 
multi-layer unsupervised learning network and a supervisor-based 
back-propagation network\cite{DBLPQuZSQ17}. Shone used asymmetric depth 
self-encoders to learn network traffic characteristics in an unsupervised, not 
only achieved good performance on large data sets, but also reducing training 
time\cite{DBLPTPS18}. Yin proposed a model using RNN, compared the 
effectiveness of the non-depth model, and achieved good 
performance\cite{DBLPZFH17}. Kim proposed a model using LSTM and gradient 
descent strategy. The experiment result which proved the LSTM can achieve a 
better performance\cite{Le2017}. Sheraz conducted a comprehensive study on deep 
learning model, and proved that the deep learning method can not only be used 
in this field, but also can achieve better performance\cite{DBLPSKBHIH18}.

\subsection{Reinforcement Learning}
Reinforcement learning is commonly considered as a general machine learning 
model, it mainly studies how agent can learn certain strategies by interacting 
with the environment, to maximize long-term reward. RL is based on the Markov 
Decision Process(MDP)\cite{DBLPSutton98}. A MDP is a tuple $(\emph{S}, 
\emph{A}, \emph{T}, \emph{R}, \gamma$), which \emph{S} is the set of states and 
\emph{A} is the set of actions. $\emph{T}$($s_i$$\vert$$s_j$, \emph{a}): 
$\emph{T}\times\emph{A}$$\to$${R}$ is the reward after executing action 
$\emph{a}$ at stage $s_i$, and ${\gamma}$ is the discounting factor. We used 
$\pi$ to denote a stochastic policy, $\pi$($\emph{s},\emph{a}$): 
$\emph{T}\times\emph{A}$$\to$[0,1] is the probability of executing action 
$\emph{a}$ at state $\emph{s}$ and $\sum_{\emph{a}\in\emph{A}}$ 
$\pi$($\emph{s},\emph{a}$)=1 for any $\emph{s}$. The goal of RL is to find a 
policy $\pi$ that maximizes the expected long-term reward. Besides, the 
state–action value function is 

\begin{equation}\label{first_equation}
\begin{array}{l}
{Q}^\pi(s,a)=E\left[ \sum_{t=0}^{\infty}{\gamma}^t 
R(s_t,a_t)|s_0=s,a_t\sim\pi(s_t) \right] 
\end{array}
\end{equation}

which $\gamma \in (0,1]$ measure the importance of future reward to current 
decisions. 

For different policies $\pi$, they represent the possibility of different 
actions selected in the same state, and also correspond to different rewards. A 
better policy can select better action in the same state, to obtain more reward.

In traditional RL, the action-value function is calculated interactively, and 
will eventually converge and obtain the optimal strategy, mainly including 
Dynamic Programming, Monte Carlo  Method and Temporal-Difference Learning. 
After deep learning was proposed, the deep reinforcement learning method formed 
by combining RL is the mainstream method at present.

In the following, we introduce the mainstream RL algorithm DDPG.

\textbf{Deep Deterministic Policy Gradient (DDPG)}: DDPG\cite{lillicrap2015} is 
a learning method that integrates deep learning neural network into 
Deterministic Policy Gradient(DPG)\cite{Silverarticle}. Compared with DPG, the 
improvement the use of neural network as a policy network and \emph{Q}-network, 
then used deep learning to train the above neural network. DDPG has four 
networks: actor current network, actor target network, critic current network 
and critic target network. In addition to the four network, DDPG also uses 
experience playback, which is used to calculate the target \emph{Q}-value. In 
DQN, we are copying the parameters of the current \emph{Q}-network directly to 
the target \emph{Q}-network, that is $\theta^{Q '}=\theta^Q$, but DDPG use the 
following update:

\begin{equation}\label{equation5}
\left  \{
\begin{array}{l}
{\theta ^{Q'}} \leftarrow \tau {\theta ^Q} + (1 - \tau ){\theta ^{Q'}}\\\\
{\theta ^{\mu '}} \leftarrow \tau {\theta ^\mu } + (1 - \tau ){\theta ^{\mu '}}
\end{array}
\right.
\end{equation}

where $\tau$ is the update coefficient, which is usually set as a small value, 
such as 0.1 or 0.01.
And this is the loss function:

\begin{equation}
L(w) =\frac{1}{m}\sum\limits_{j=1}^m(y_j-Q(\phi(S_j),A_j,w))^2
\end{equation}

\section{Methods}

\begin{figure*}[h]
	
	\centering
	\includegraphics[width=16.8cm]{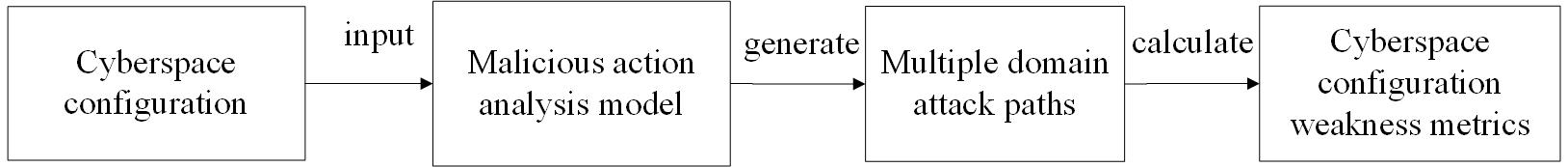}
	
	\caption{Cyberspace Configuration Weakness Analysis Architecture}
	
	\label{figure11}
	
\end{figure*}

The influence of cyberspace security is mainly reflected in different 
cyberspace configurations, which can influence the length or concealment of 
attack paths in cyberspace. Different configurations in cyberspace can change 
attacker's attack paths in different ways. It is generally believed that 
cyberspace contains physical domain, digital domain, cognitive domain, social 
domain, etc. Since the relationship between people and people in the social 
domain generally changes in real time, it is not considered for the time being. 
In this paper, the relevant configuration and related authority of physical 
domain, digital domain and cognitive domain are mainly considered.

After analyzing the relation between cyberspace authority, namely can analysis 
the cyberspace configuration weakness, its basic idea is if an attacker can 
find attack paths under current cyberspace configuration. If under a certain 
cyberspace configuration, the attacker can more easily find attack paths, as a 
result, the attacker can easily modify the cyberspace configuration and obtain 
the security information, illustrates the cyberspace configuration is bad.
On the other hand, the security of cyberspace configuration is necessary. So we 
proposed the method to analysis the weakness of cyberspace, to find the attack 
paths to analysis weakness in the current cyberspace, then enhance the 
cyberspace security based the weakness analysis result.

\subsection{The Weakness Analysis Architecture}

The input of cyberspace configuration weakness analysis architecture is the 
current cyberspace configuration, through the malicious action analysis model 
to analysis its possible malicious actions, get the attacker's attack paths, 
ultimately rely on the cyberspace multiple domain attack paths, according to 
the cyberspace configuration index value, calculate the weakness of the current 
cyberspace configuration. The architecture is shown in Figure \ref{figure11}.

This process is divided into two core processes, first is through the current 
cyberspace configuration, we will get the corresponding cyberspace attack paths 
based on the malicious action analysis model, this is the most critical in this 
process. In this process, we use the DDPG algorithm, and represent an attacker 
as an agent. The agent can learning cyberspace intrusion policy by its 
autonomous, and then find the attack paths in cyberspace, therefore the agent 
can find the hidden attack paths. Second is calculating the cyberspace 
configuration's weakness, mainly through compute cyberspace attack paths to 
calculate cyberspace configuration weakness value. In the process, because of 
the different attackers' initial permissions, the cyberspace attack paths is 
different, so we need to comprehensive analysis of different attackers, 
integrated to the cyberspace attack path of different initial permission 
attackers, and corresponding the weakness of cyberspace configuration is 
obtained.

\subsection{Malicious Action Analysis Model}

The main function of the malicious action analysis model is to obtain the 
possible cyberspace attack paths of an attacker who given its initial 
permission under a certain cyberspace configuration. In this process, we use 
DDPG algorithm, and the attacker with initial permissions are represented as 
the agent, who can learn cyberspace intrusion strategy autonomously and obtain 
cyberspace attack paths.

\begin{figure}[h]
	
	\centering
	\includegraphics[height=4.2cm, width=6.8cm]{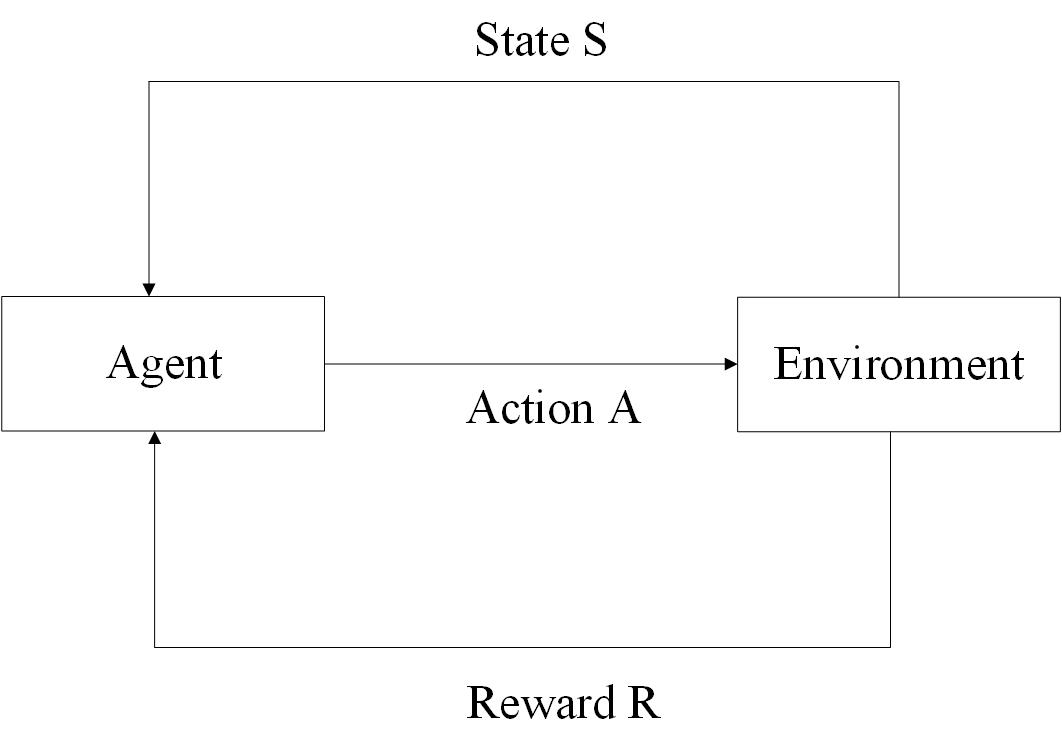}
	
	\caption{The Malicious Action Analysis Model}
	
	\label{figure22}
	
\end{figure}

Specifically, the malicious action analysis model takes DDPG algorithm, we are 
using an agent to represent the likely attacker. In the process of finding the 
attack paths, the agent first selects the action in the current state, which 
can change the environment(cyberspace configuration) and the agent's state. At 
the same time the agent will obtain certain reward, negative reward(captured by 
administrator) or none. Besides, the change in the agent's state enables it to 
perform other actions to obtain more rewards. As a result, the agent finds 
attack paths in this cyberspace configuration by trial and error. The model is 
shown in Figure \ref{figure22}.

\begin{figure*}[h]
	
	\centering
	\includegraphics[width=15.2cm]{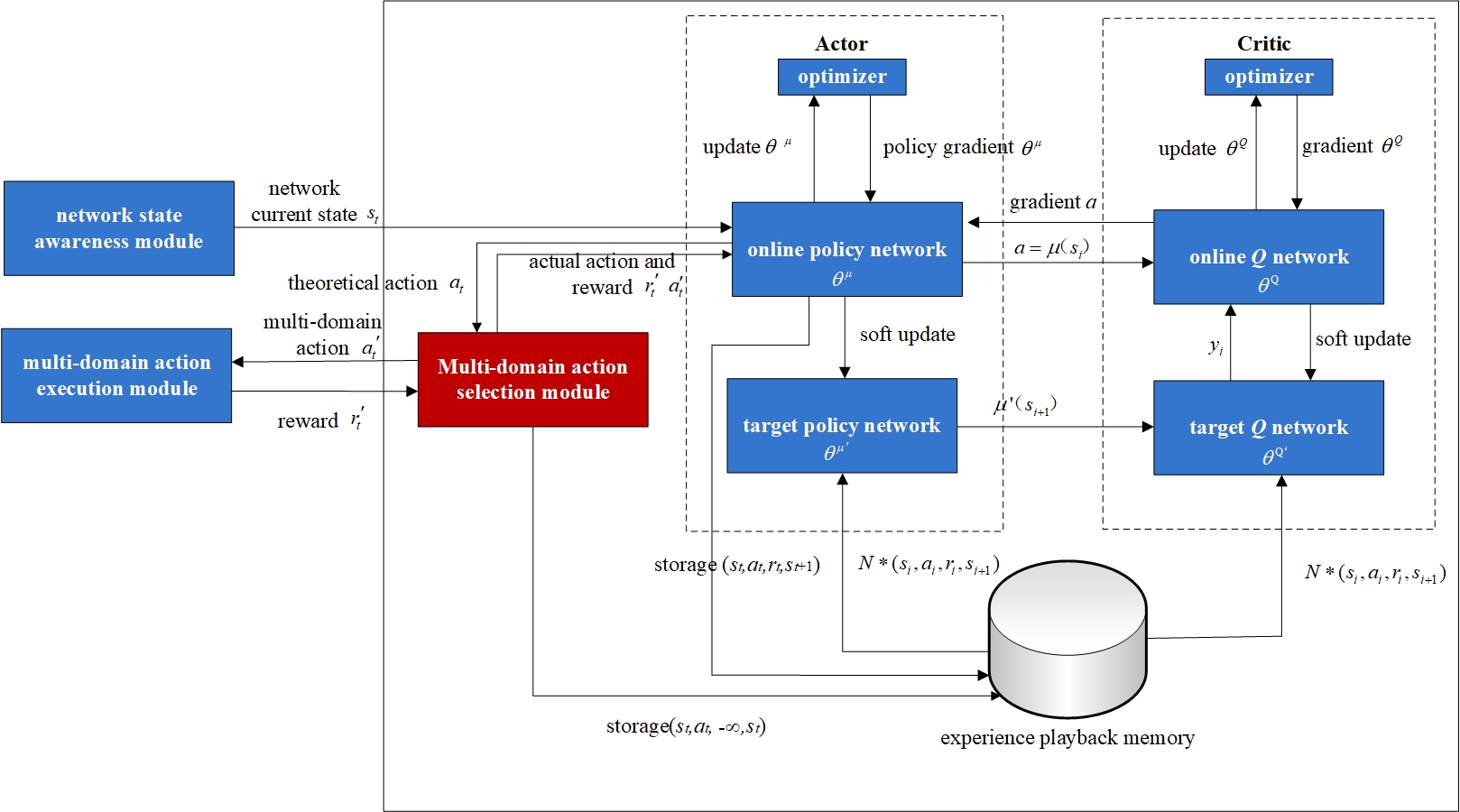}
	
	\caption{The Improved DDPG Architecture}
	
	\label{figure33}
	
\end{figure*}
The goal of malicious action analysis model is to find an optimal policy given 
a cyberspace state, and to select the corresponding action \textbf{\emph{a}} 
according to the current state \textbf{\emph{s}} of the cyberspace, it also 
means find the corresponding policy mapping function $R(s) \to A$, to make the 
long-term reward of agent maximum. In this process, the policy can be divided 
into two categories, namely the deterministic policy and the stochastic policy, 
deterministic policy is for the state, the conviction of corresponding 
output(action). In general, the deterministic policy algorithm efficiency is 
high, but the lack of ability to explore and improve. Rather than the 
stochastic policy is based on deterministic policy, to join corresponding 
random value, enables the stochastic policy to have certain ability of 
exploration. For the malicious action analysis model, since the action value 
range is generally not large in practical problems, a deterministic policy is 
adopted to ensure better performance of the model.

The malicious action analysis model treats the problem as a MDP, that is $M = 
(S,A,P,R,\gamma )$, where is $s \in S$  the current state of the cyberspace,  
$a \in A$ is an attack action that is currently available, $P$ is the 
probability of transitions between states, $R$ is the reward value after taking 
an action to reach the next state. $\gamma$ is the discount factor. For the 
transfer probability, it can be expressed formally as $p(\hat s|s,a) = p({S_{t 
+ 1}} = \hat s|{S_t} = s,{A_t} = a)$. For the reward function, it can be 
formally expressed as $R(s,a) = E[{R_{t + 1}}|s,a]$.

On the specifically model, the malicious action analysis model is a standard RL 
model, through the study of the awareness of environment, the agent will take 
the action and get a reward, the goal of the agent is to maximize rewards, and 
then to further training of the agent. In this paper, we will to take DDPG 
algorithm, and its main architecture as shown in Figure \ref{figure33}.

With the standard DDPG algorithm, the basic architecture of the malicious 
action analysis model also consists of four networks and one experience replay 
memory. Among them, the experience replay memory is mainly responsible for 
storing the state transfer process of $ <s,a,r,s'> $, then, by means of small 
batch sampling, the corresponding transferred samples are extracted to train 
the corresponding neural network so as to avoid the strong correlation between 
samples. Among the four networks, there are two policy networks(Actor) and two 
\emph{Q}-networks(Critic), namely the online policy network, the target policy 
network, the online \emph{Q}-network and the target \emph{Q}-network. The 
policy network mainly simulates the attacker's policy through the deep neural 
network, which takes the current state as the input and the output as the 
corresponding action. The \emph{Q}-network is mainly used to estimate the 
expectation of the final reward value obtained if the policy is continuously 
executed after the current action is executed in a certain state. The input is 
the current state, the current action and the output is the \emph{Q}-value. If 
only a single neural network is used to simulate policy or \emph{Q}-value, the 
learning process is unstable. So in DDPG algorithm respectively, policy network 
and \emph{Q}-value network create copies of two networks, two networks are 
known as the online network, two networks are known as the target network, 
online network is the current training of network, the target network is used 
to calculate the training goal, and after a short period of time, the model of 
online networks parameter updates to the target networks, so as to make the 
training process is stable, easy to convergence.

We have improved the standard DDPG algorithm, which is different from the 
standard DDPG algorithm in three aspects:

\begin{itemize}
	\item  In improved DDPG algorithm, we introduced the multi-domain action 
	selection module.
	
	Different from standard DDPG algorithm, the biggest change is that the 
	introduction of multi-domain action selection module. In the standard DDPG, 
	the actions which agent can choose in each state is the same. But in this 
	environment, when the attacker select the attack paths in cyberspace, he 
	have different alternative actions in each state. In order to make DDPG 
	algorithm can choose different actions in different states, joined the 
	multi-domain action selection module. This module's input is online policy 
	network's output, theory action $a_t$, then a linear change under this 
	current state, and perform the actual action $a_t'$, the actual execution 
	action $a_t'$ into multi-domain action execution module, get the 
	corresponding reward $r_t'$. In the end, the corresponding actual execution 
	of action $a_t'$ and the corresponding reward  $r_t'$ return online policy 
	network. Through this method, the reasonable choice of actions in different 
	states can be realized.
	
	\item Second, the input of experience playback memory is different.
	
	In order to ensure that the multi-domain action selection conforms to the 
	constraints of the actions on the state, the input of the experience 
	playback memory is increased, not only by the online policy network to 
	store the sequence $({s_t},{a_t},{r_t},{s_{t + 1}})$, that is, execute the 
	action $a_t$ in the state$s_t$, get the reward value $r_t$, and convert the 
	relevant state to $s_{t+1}$. Moreover, since the corresponding relationship 
	between the state and the action needs to be considered when selecting the 
	action, it is avoided that the policy network chooses the action that is 
	not feasible in the state. Therefore, when the policy network selects an 
	inoperable action $a_t$ in the state $s_t$, it is not only necessary for 
	the multi-domain action selection module to use a linear transformation to 
	map it to a feasible action $a_t'$. In addition, relevant action sequences 
	$({s_t},{a_t},{- \infty},{s_{t + 1}})$ need to be taken to indicate that 
	actions $a_t$ are executed in the state $s_t$, and the subsequent state 
	obtained is still $s_t$, and the reward at this time is a huge negative 
	value, so as to ensure that relevant actions are not selected in the 
	process of training the policy network.
	
	\item At last, the architecture of the policy network is changed based on 
	the relevance of the input state. 
	
	In terms of network architecture, the two policy networks have the same 
	architecture, whose input is the state of network and output is the action 
	to be selected. Structurally, a RNN hidden node is added between the 
	original DDPG input layer and the hidden layer. The transformed policy 
	network is divided into 5 layers. The first layer is the input layer; The 
	second layer is the RNN hidden layer, which contains 32 GRU nodes. Layer 
	3th and layer 4th are the full connected layer, including 48 full connected 
	nodes. The activation function uses the ReLu function. The fifth layer is 
	the output layer, use the sigmoid function as the activation function, and 
	finally output a multi-dimensional vector representing the multi-domain 
	action that needs to be performed. 
\end{itemize}

In addition, the two \emph{Q}-networks have another architecture, whose input 
is not only the state of the network, but also includes a multidimensional 
vector, representing the corresponding multi-domain actions, and the output is 
a scalar, representing the corresponding \emph{Q}-value of the corresponding 
states and actions. The network is divided into four layers. The first layer is 
the input layer; The second layer and the third layer respectively contain 48 
fully connected nodes. The activation function uses the ReLu function. The 
fourth layer is the output layer, which outputs a scalar and uses the linear 
function as the activation function, representing the corresponding 
\emph{Q}-value of the corresponding state and action.

\subsection{Cyberspace Configuration Weakness Metrics}

The measurement of cyberspace configuration weakness is the basic index 
measured by the average attack path of all users in the cyberspace, as shown in 
equation \ref{equ4}:

\begin{equation}\label{equ4}
sec (s) = \mathop {\lim }\limits_{n \to \infty } \frac{{\sum\limits_{i = 1}^n 
{len(A({u_i}))} }}{n}
\end{equation}

Among them, \emph{s} is the multiple domain configuration of the current 
cyberspace, which is the object to be evaluated; $sec(s)$ is the security 
measures configured for the current multiple domain cyberspace; $n$ is the 
number of attackers. For the same attackers with different initial states, they 
can be considered as different network attackers, $u_i$ is the user $i$, 
$A(u_i)$ is the shortest attack path of the user $i$, $len(path)$ is the length 
of the path $path$. The attack path is the attacker how he can get the security 
information from the initial permission through the relevant steps.

By the equation \ref{equ4} can be seen that for measurement cyberspace 
configuration weakness, can be turned into search for the most likely attack 
paths, by this measure, the cyberspace configuration weakness metric into 
intelligence agent to autonomous learning, to enhance the automation of the 
cyberspace security configuration has the profound significance.

\section{Experiments}

\subsection{Experiment Environment}

\begin{figure}[h]
	
	\centering
	\includegraphics[height=4.2cm, width=6.8cm]{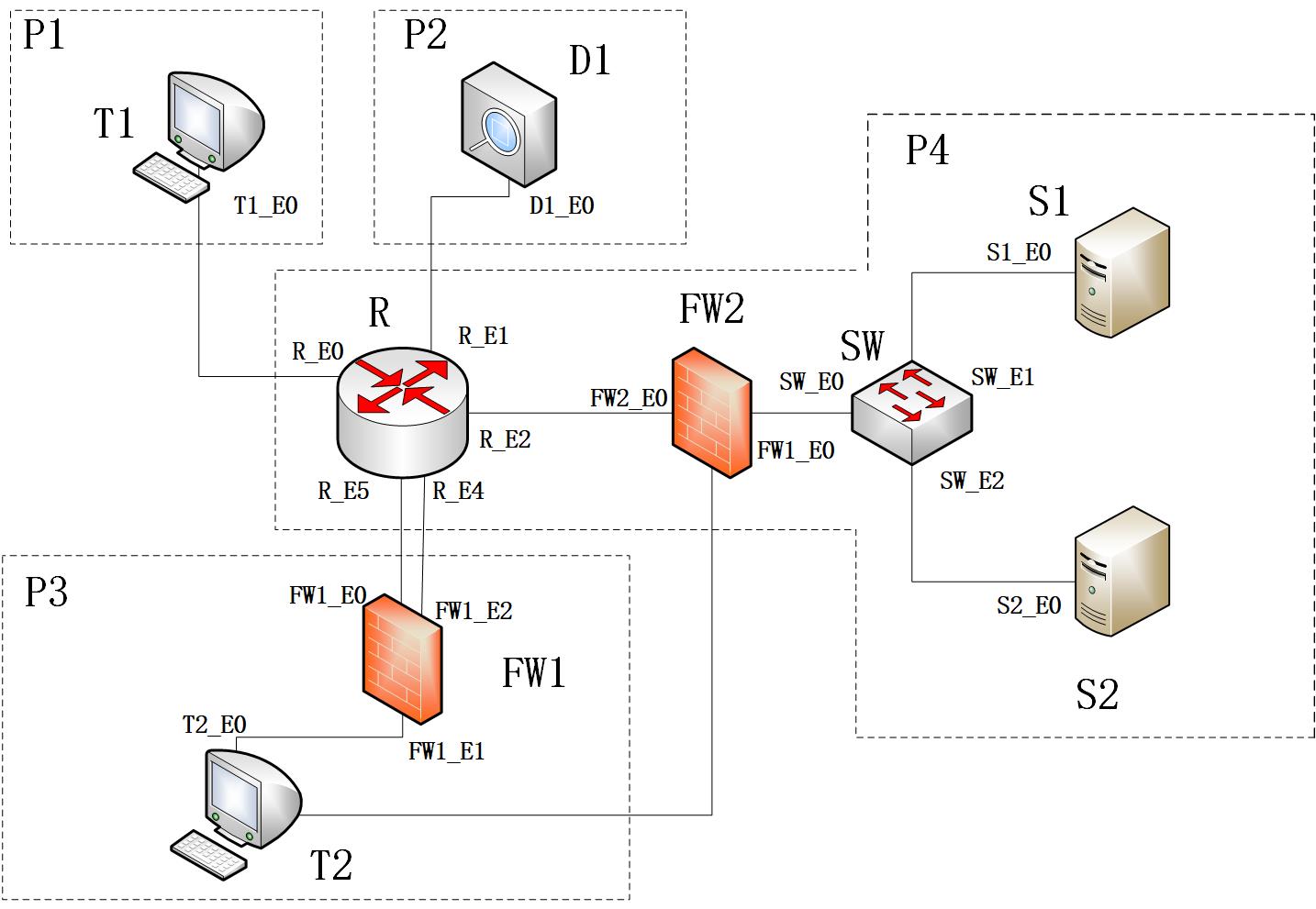}
	
	\caption{Experiment Environment Topology}
	
	\label{figure336}
	
\end{figure}

In the experiment, the corresponding simulation environment is constructed to 
verify the effectiveness of the method. In this environment, there are five 
spaces in total. The outermost space is the whole physical space, representing 
a region. P1 is the region where terminal located, P2 is the region where VPN 
equipment located, P3 is the location of the communication team, and P4 is the 
communication hub. There are 5 kinds of equipment, including computer 2 sets 
(T1 and T2, respectively stored in P1 and P3), firewall 2 sets (FW1, FW2, 
respectively stored in P3 and P4), sensor (D1, stored in P2), router (R, stored 
in P4) and switch (SW, stored in P4), server 2 sets (S1, S2, stored in P4) and 
its equipment connection relationship as shown in Figure \ref{figure336}. The 
security information is stored in S2.


In this network, there are 15 network services, as shown in Table \ref{tab111}

\begin{table*}
	\caption{Network Services in the Experiment Environment}
	\label{tab111}
	\begin{tabular}{ccccc}
		\toprule
		Web Service & Web Services's Role & Service Support Equipment & Service 
		Dependent Port & Service Password\\
		\midrule
		T2\_manager & Remote Management equipment & T2 & T2\_E0 & None\\
		FW1\_manager & Remote management equipment & FW1 & FW1\_E2 & 
		FW1\_password\\
		FW2\_manager & Remote management equipment & FW2 & FW2\_E2 & 
		FW2\_password\\
		S1\_web & Web services in server S1 & S1 & S1\_E0 & S1\_web\_password\\
		S2\_web & Web services in server S2 & S2 & S2\_E0 & S2\_web\_password\\
		\bottomrule
	\end{tabular}
\end{table*}

In this environment, because of the firewall FW1 equipment are in need of 
remote management, FW1-password remains in FW1, at the same time, due to the T2 
maintains FW2 and S1, so T2 store password FW2\_password and S1\_web\_password, 
in this environment, to ban other flow of information. But we can know, an 
attacker can through the multiple domain joint attack, which can obtain the 
security information stored on the server S2, a possible attack path is as 
follows:

First, attacker enter space P2 and obtain the management service password 
FW1\_password of firewall FW1;

Second, use device T1 or D1 to access the management service of FW1, add access 
control list: allow T1 or D1 to access the management service of T2, that is 
T2\_manager;

Third, get the password FW2\_password of firewall FW2 stored on T2 and the 
password S1\_web\_password of service S1\_web through T2\_manager;

Fourth, use T2\_S1 port, access firewall FW2\_manager, add access control list: 
allow T1 or D1 access service S1\_web and S2\_web;

Fifth, use T1 or D1 to access the service S1\_web and get the password 
S2\_web\_password of S2\_web, at this point, the attacker's higher permissions 
have been obtained.

At last, the attacker can use T1 or D1 to access the service S2\_web to get the 
security information by the S2\_web\_password.

In this process, three key firewall security policy changes are involved: on 
firewall FW1, T1 or D1 are allowed to access T2's management service 
T2\_manager; On firewall FW2, allow T1 or D1 access to service S1\_web; On 
firewall FW2, allow T1 or D1 access to the service S2\_web.

\subsection{Experiment Process}

During the experiment, an agent(attacker) is introduced, located in the 
outermost space, and then, in the environment shown in Figure \ref{figure336}, 
for three key security policies (on firewall FW1, T1 or D1 are allowed to 
access T2's management service T2\_manager; On firewall FW2, allow T1 or D1 to 
access service S1\_web; On firewall FW2, allow T1 or D1 to access the service 
S2\_web). Randomly add 0 or more security policies, and respectively calculate 
the average attack action length of the attacker to obtain security information 
when the number of key security policies is different (if the attack action 
length exceeds 10000, it will be forced to quit, which means that the attack is 
unsuccessful, otherwise, the action sequence length of the first time to obtain 
the security information will be recorded).

According to the malicious action analysis model, according to the DDPG 
algorithm, define the corresponding state, action, reward, etc. The relevant 
settings are as follows:

On the set of state, with a length of 106 vector to represent a state of 
different position on the value of the vector may users, respectively from the 
spaces, the ports, services or information, in setting a state vector, if in 
the state, the attacker exists, will be to the attacker's value is set to 1, 
otherwise 0. If the attacker is in a certain space, the value representing that 
space is set to 1; otherwise, it is 0. If the attacker uses a port, set the 
value representing the port to 1, otherwise 0; If the attacker is connected to 
a service, set the value representing the service to 1, otherwise 0; If the 
attacker obtain security information, set the value representing that 
information to 1, otherwise 0.

Attackers have different action in different states. For example, when the 
attacker can dominate the management service FW1\_manage of FW1 in the current 
state, he can add the corresponding access control list for firewall FW1 in the 
current state. Otherwise, he cannot add the access control list for FW1. For 
example, if the attacker is able to access the service S1\_web in the current 
state and has the password S1\_web\_password, he can dominate the service. If 
it has access to the service S1\_web, but does not have the password 
S1\_web\_password, he cannot dominate the service.

In terms of the reward setting, different reward are set for the attacker 
according to the degree of the completion of the attack path. Among them, when 
the attacker can dominate the management service FW1\_manage of FW1, its 
current reward is 100. When the attacker can dominate the management service 
FW2\_manage of F2, its current state reward is 200. When the attacker obtains 
the administrative service password S1\_web\_password of the S1\_web, the 
reward is set to 300. When the attacker obtains the administrative service 
password S2\_web\_password of S2\_web, the reward is set to 400. When the 
attacker obtains the final security message, the reward is set to 10000.
\begin{figure}[h]
	\centering
	\subfigure[]{		
		
		{\includegraphics[width=6.9cm]{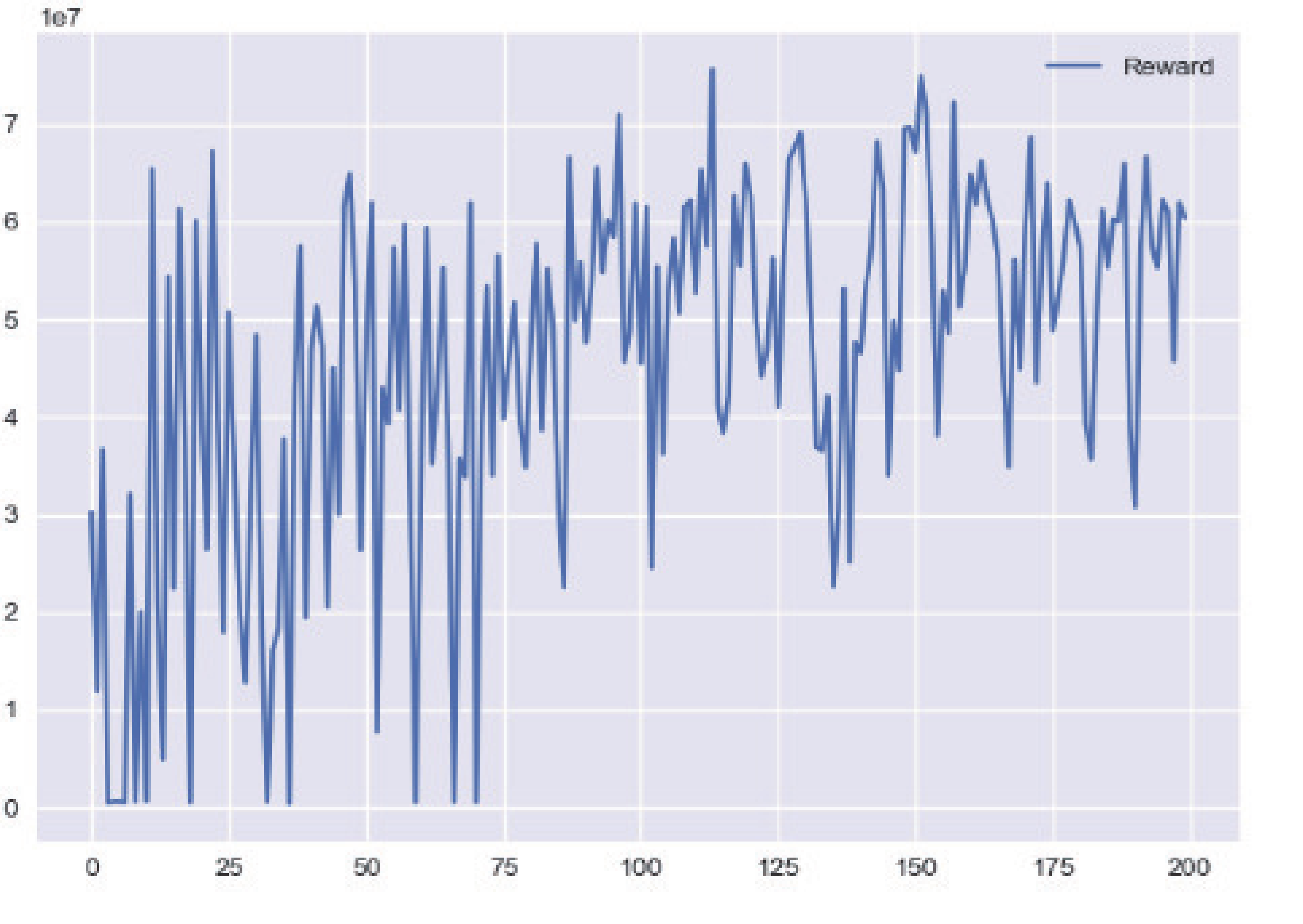}}
		\label{ExpResult1}
	}
	
	\subfigure[]{	
		{	\includegraphics[width=7.6cm]{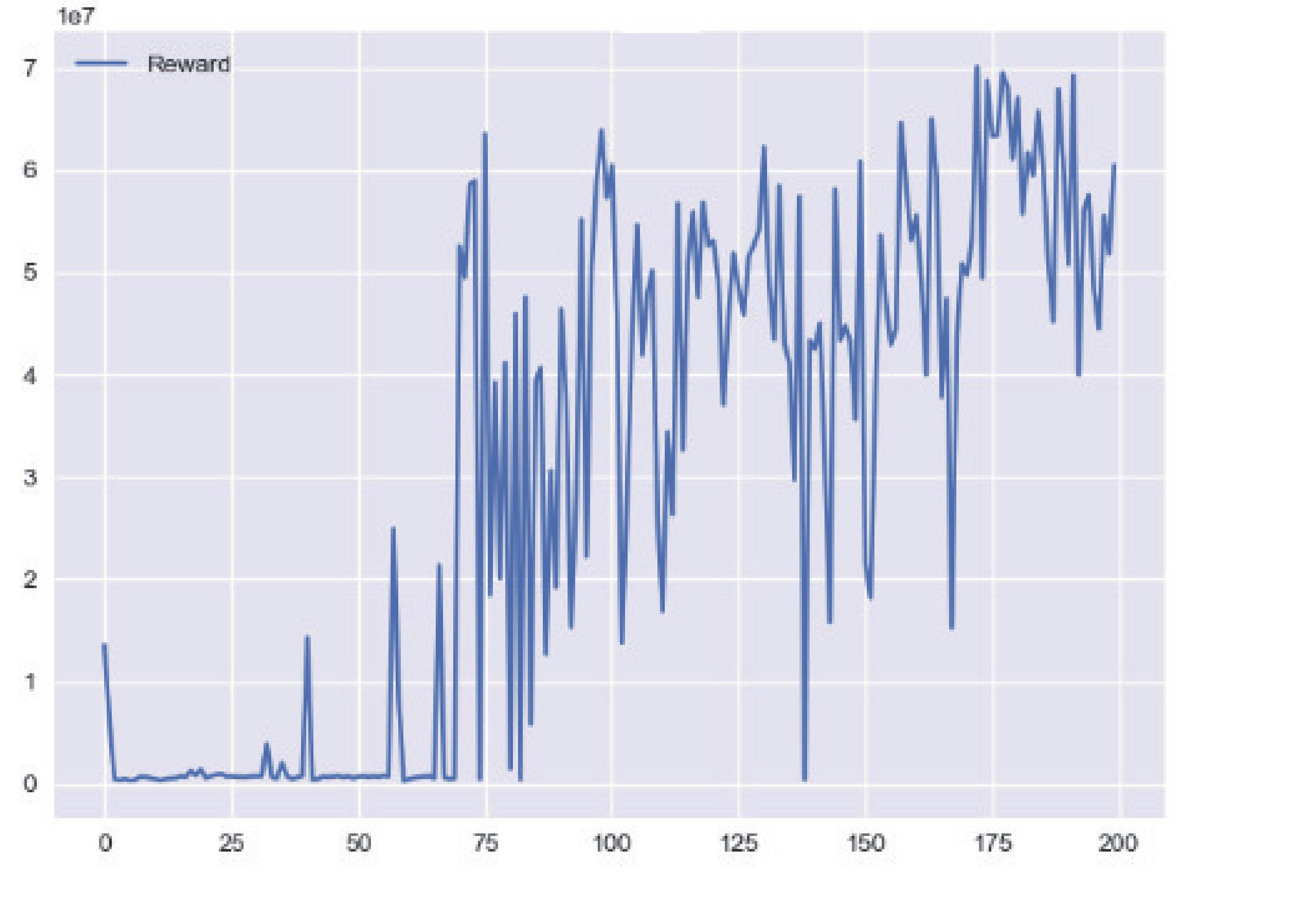}}
	}	
	\caption{Attacker's Reward Value Change in Training}	
	\label{ExpResult2}	
\end{figure}
\begin{figure}[h]
	
	\centering
	\includegraphics[height=4.2cm, width=6.8cm]{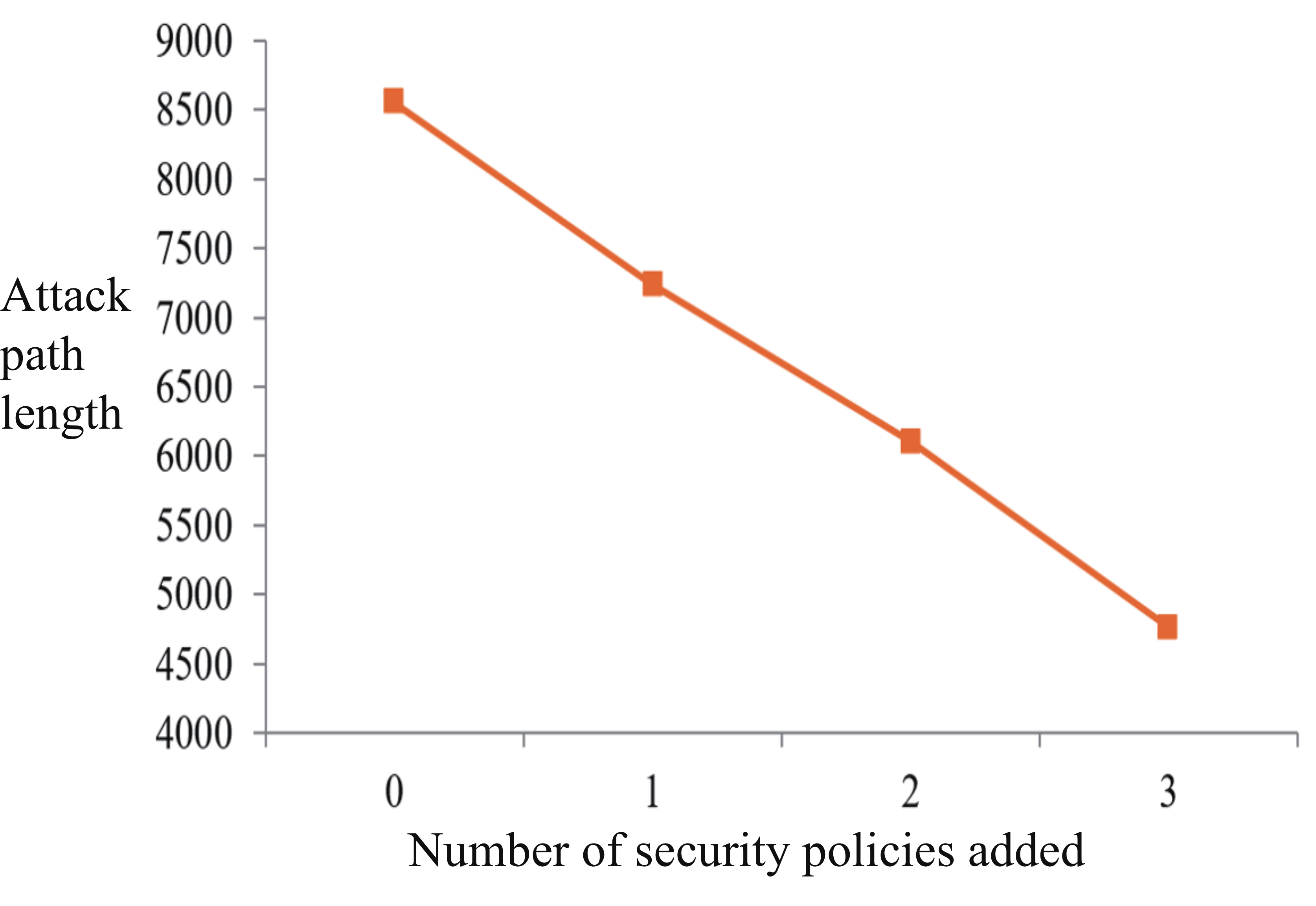}
	
	\caption{Length of Attack Path in Different Firewall Security Policy}
	
	\label{ExpResult3}
	
\end{figure}

\subsection{Results and Discussions}

Under the above conditions, learn 60 times in different environments 
respectively, and calculate the corresponding attack sequence length. First of 
all, the attacks on a policy for statistical learning process, the process of 
two typical as shown in Figure \ref{ExpResult2}, with the increasing the number 
of training, \emph{R} present a slow upward trend, until finally tend to be 
convergent, it is in a learning process of RL, shows that the proposed model 
can monitor to the attacker's action of gradually learning the characteristics 
of a attacker's action rule, and constantly improve its accuracy of judgment, 
so as to verify the effectiveness of the proposed model is presented in this 
paper.

Secondly, the length of the attack path under different security configurations 
is calculated. According to the number of policies added, the length of the 
possible attack path is calculated, as shown in Figure \ref{ExpResult3}.

From Figure \ref{ExpResult3}, adding in the key security policies, the more 
attacks the attackers more easily to the anticipated target, this also from the 
side the important of the configure security. Besides, the length of the 
attacker's attack path in this configuration is used to measure the weakness of 
cyberspace, it also shows that the level of cyberspace configuration security 
can affect the attacker attack difficulty, this verified the correctness of the 
proposed method.

\section{Conclusion}
Based on the current cyberspace configuration lack of multiple domains attack 
evaluation, we proposed the weakness analysis of cyberspace configuration based 
on reinforcement learning. Meanwhile, we has been learn about the cyberspace 
weakness metrics, and finally has carried on the experimental verification. 
This method can comprehensively consider the mutual influence of the multiple 
domain configuration in the cyberspace, and can take an intelligent method to 
analysis the weakness of the cyberspace, which has a strong practical value.

This paper analysis a typical cyberspace environment and applies the 
reinforcement learning method to analysis of cyberspace configuration, which 
has achieved better results. However, the cyberspace environment in this paper 
is limited. In the next step, we hope to apply the reinforcement learning to 
more cyberspace operation and maintenance management, and achieve better 
results.


\bibliography{example}
\bibliographystyle{ACM-Reference-Format}
\end{document}